# AN EFFICIENT DEPENDENCY PARSER USING HYBRID APPROACH FOR TAMIL LANGUAGE

K.Sureka
Student,Dept. of CSE-PG,
National Engineering College,
Kovilpatti, Tamilnadu,India.
surekakrishcs@rediffmail.com

Dr.K.G.Srinivasagan
Prof. & Head, Dept. of CSE-PG,
National Engineering College,
Kovilpatti, Tamilnadu,India.
kgsnec@rediffmail.com

S.Suganth
Asst.Prof. Dept. of CSE-PG,
National Engineering College,
Kovilpatti, Tamilnadu,India.
Krish.sugi1@gmail.com

*Abstract*—Natural language processing is a prompt research area across the country. Parsing is one of the very crucial tool in language analysis system which aims to forecast the structural relationship among the words in a given sentence. Many researchers have already developed so many language tools but the accuracy is not meet out the human expectation level, thus the research is still exists. Machine translation is one of the major application area under Natural Language Processing. While translation between one language to another language, the structure identification of a sentence play a key role. This paper introduces the hybrid way to solve the identification of relationship among the given words in a sentence. In existing system is implemented using rule based approach, which is not suited in huge amount of data. The machine learning approaches is suitable for handle larger amount of data and also to get better accuracy via learning and training the system. The proposed approach takes a Tamil sentence as an input and produce the result of a dependency relation as a tree like structure using hybrid approach. This proposed tool is very helpful for researchers and act as an odd-on improve the quality of existing approaches.

## I. INTRODUCTION

Natural Language Processing (NLP) is an area of research and application that explores how computers can be used to understand and manipulate natural language text or speech to do useful things. The Origin of Natural Language Processing lie in a number of disciplines such as, computer and information sciences, linguistics, mathematics, electrical and electronic engineering, artificial intelligence and robotics, psychology, etc. Machine translation is one of the major application area under language processing. The ideal aim of machine translation systems is to produce the best possible translation without human aid. The structural order might vary from language to language. During the translation of English-Tamil, the structural order might be a difficult task because the order of words may affect the original meaning of a sentence. In order to translate correct interpretation of a sentence, the word structure order is very essential. The syntactic rules are used to predict the correct sentence translation based on the target language structure and these are also used to improve the optimization of translation process. The machine translation requires the parsed output to be translated. Parsing gives the structural analysis of the sentence. There exist tools like Stanford parser which gives the dependency information about the sentence. But, the tool gives the dependency structure only for the English. There are no such efficient tools for the Tamil language. Dependency parsing suits the best to get the structural information about the sentence.

Dependency parsing is a very useful tool in the sentence structure identification. For languages like English, it can be easily told that the sentence structure is in the form of SVO (Subject Verb Object). Though the sentence structure for the Tamil is SOV (Subject Object Verb) form, it is not necessary that the sentences should follow the same format as the target language. Generally Tamil language can accept to write the sentence in any pattern because Tamil sentence structure is changed, then the exact meaning of the sentence also is changed. So, in these cases of free word order language, dependency parsing gives the correct sentence structure.

The clause boundary identification is a vital task in language generation and understanding. Parsing gives the best way to solve the clause boundary identification. If the source sentence length is too long, it can be split based on the clauses which make the translation very easy. The clauses in the sentences may be joined using the connectives or the clauses may be embedded within the other clauses. Sometimes, the two clauses are separated only by commas, in those cases, the translation is easy. Identifying the clauses in the sentences itself becomes a complex task. Dependency parsing gives the information about the relation extraction as how the words in the sentence are related and what type of relationship exists between the words in the sentence. So, considering the various uses of the parsing, an efficient tool needs to be developed.

In proposed work, a hybrid approach is proposed that uses both techniques i.e. rule based and machine learning to build an identifier for different clause boundaries of Tamil language. The POS tagger and Chunker are used to prepare the parts of speech and chunked tagged data as the inputs, where linguistic rules are taken as features. The rest of the paper is organized as follows, the related works are discussed in section 2, and framework and proposed algorithm presented in section 3, implementation methodology in section 4, experimental results and discussion are reported in section 5 and followed by concluding remarks.

## II. RELATED WORK

The researchers are reported the different levels of accuracy as result. The results may not fulfill the scenario of users need various dependency parsing works has been experimented for different languages. Some of the work has been focused as follows.

The Tamil Shallow Parser was developed using the new and state of the art machine learning approach [1]. The Shallow Parser system developed for Tamil is an important tool for Machine Translation between Tamil and other languages. A shallow parsing approach is compared with PP-attachment [2] with a state of the art full parser. It is used a flat representation of prepositional phrases and their associated attachment sites to train a machine learner for the PP attachment task. A memory-based approach can obtain results for the PP attachment task comparable to a state-of-the art full parser. A phrase structured Treebank has been developed with 326 Tamil sentences which covers more than 5000 words. A hybrid language model has been trained with the phrase structured Treebank using immediate head parsing technique. Lexicalized and statistical parser which employs this hybrid language model and immediate head parsing technique gives better results than pure grammar and trigram based model [3].

The data-driven dependency parsing [4] have shown that the distribution of parsing errors are correlated with theoretical properties of the models used for learning and inference. This experimental results show that both models consistently improve their accuracy when given access to features generated by the other model, which leads to a significant advancement of the state of the art in data-driven dependency parsing. The clause markers play the role to detect the type of sub-ordinate clause, which is with or within the main clause. Limitation with CRFs is that it is highly dependent on linguistic rules. Missing of these rules may lead to wrongly classified data [5].

## III. FRAME WORK OF DEPENDENCY TAMIL PARSER

The implementation of dependency parsing involves a sequence of several steps.

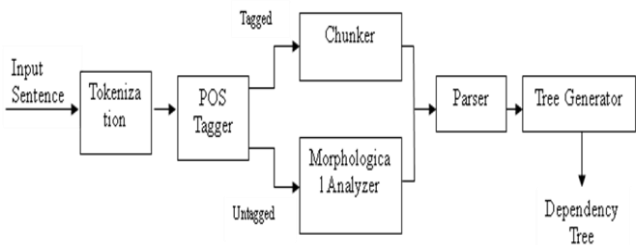

**Fig 1. Framework of Dependency Parser**

The input sentence is initially be tokenized, the tokenized words are sent it to the pos tagger to get the grammatical label of each data. It is the process of automatically assigning the label based on the grammatical category or lexical class to each and every word in a sentence. It is considered to be important process in speech synthesis, speech recognition, natural language processing, information retrieval and machine translation, etc. During POS tagging, there is a possibility for ambiguity because the same word has different meaning in different contexts. It includes verbs, nouns, adjectives, adverbs, and determiner and so on. The Fig 2 represents the Parts of speech tagging.

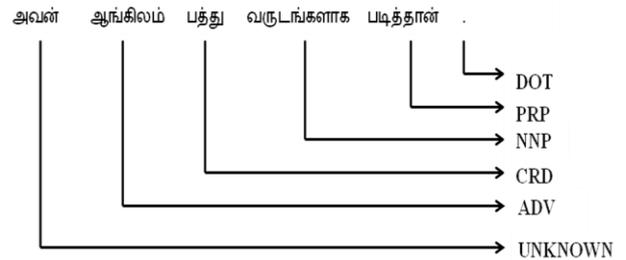

**Fig 2. POS Tagging**

Some of the words are formed in untagged. The untagged words are passed to the morphological analyzer to get the root word with appropriate tagged data. Tamil language is morphologically rich and agglutinative. Such a morphologically rich language needs deep analysis on the word level to capture the meaning of the word from its morphemes and its categories. Each root is affixed with several morphemes to generate a word. In general, Tamil language is postpositionally inflected to the root word. The Fig 3 represents the Morphological analysis of each untagged word in the parts of speech. Tamil is one of the classical Indian languages which has very strong linguistic base with well defined set of morpho syntactic rules. Generally suffixes are used to mark class, numerals and cases attached to noun or verb root.

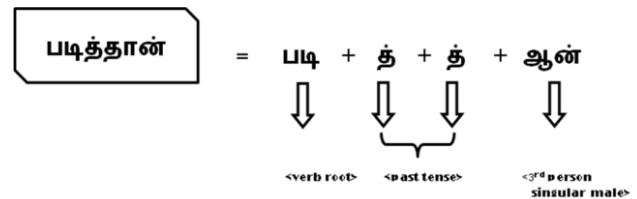

**Fig 3. Morphological Analyzer**

Tamil is one of the classical Indian languages which has very strong linguistic base with well defined set of morpho syntactic rules. Generally suffixes are used to mark class, numerals and cases attached to noun or verb root.
Chunking is an efficient and robust method for identifying short phrases in text, or "chunks". Chunking is considered as an intermediate step towards full parsing. A chunker finds adjacent, non-overlapping spans of related tokens and groups them together into chunks. Chunkers often operate on tagged texts, and use the tags to make chunking decisions. A subsequent step after tagging focuses on the identification of basic structural relations between groups of words. This is usually referred to as phrase chunking. It segments of a sentence with syntactic constituents such as noun or verb phrase (NP or VP). Each word is assigned only one unique tag, often encoded as a begin-chunk (e.g. B-NP) or inside-chunk tag (e.g. INP) and outside-chunk tag(e.g. ONP).

There is no standard sentence structure for tamil language . The sentence structure are obtained based on this grammar rules.

NP -> (Det) (Adj) **N** (PP)
VP -> **V** (NP) (PP) (Adv)
PP -> **P** (NP)
ADJP -> (CRD) (ADJ)
ADVP -> (ADV) (INT) (CRD)
NP ->  (NP) conj (NP)
S  -> (NP)* (VP)

The tagged sentence is passed to the chunker process, this is used to get the chunked data. The Fig 4 represents the group the words in the sentence.

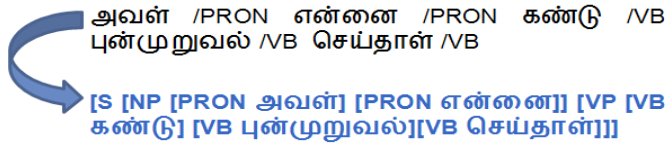

**Fig 4. Chunking**

At present the words includes the pos tag and chunk tag of each word.

Dependency parsing assigns head-dependent relations between the words in a sentence. Whenever two words are connected by a dependency relation, that one of them is the head and the other is the dependent, and that there is a link connecting them. In general, the dependent is in the form of modifier, object or complement. The head plays the larger role in determining the behavior of the pair. In our dependency representation the source of the edge represents the modifier and destination points to the head word. Fig 5 represents dependency relationship between words in the sentence. Here படித்தான் is the root word and it is the head of the other words. And the remaining word indicates dependent on head.

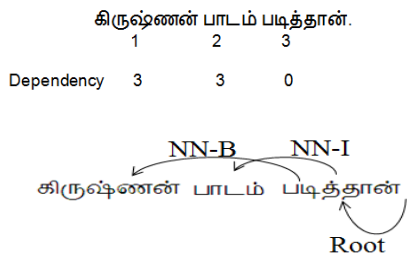

**Fig 5. Dependency Relationship structure**

This project mainly focused to create a parser tool based on the dependency priority of a word. The Dependency Parsing assigns head-dependency relations between them. The Root word of the tree is "படித்தான்". And "கிருஷ்ணன்","பாடம்",are dependent on root word. The result of dependency priority is to directly to make a manual tree structure of a sentence. In a computational way, the tree viewer is used to bring manual tree and to remake the computationally graphics structure of a words in a sentence. This resultant tree format is useful to identify the grammatical relationship of the words and also to make doing further research in a language. The Following Fig 7 illustrates the Parse Tree structure

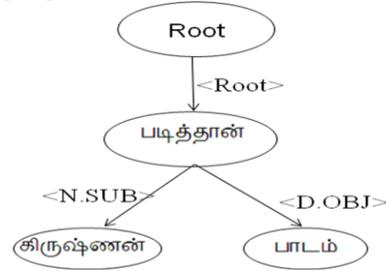

**Fig 7. Parse Tree**

### IV. HYBRID APPROACH

The Parser System using hybrid approach this approach combine with Rule based and Machine learning approaches. The Machine learning approach used for conditional random field (CRF). To achieve a better morphological analyzer using "Rule Based Forward Algorithm**"**

**Algorithm:**
**Input:**
Data Set D={($W_1$,$P_1$), ($W_2$,$P_2$)……($W_n$,$P_n$)};
**Process:**
Let the sentence be "S" and split into "$W_1$" "$W_2$"………… "$W_n$";
For i=1 to n;
{$W_1$ ,$W_2$………… $W_n$ } _ { ({$W_i$ }If({$W_i$ }_{D}) then
Goto Stemming:
**Stemming:**
Let {$W_i$ } as a string;

Root Word R;

Verb Root VBR;

Noun Root NR;

Take {$W_i$ } split into {$C_1$, $C_2$…….. $C_n$ }

ArrayList Ai;

Store {$C_1$, $C_2$…….. $C_n$ } into $A_i$ ;

Use Pattern Match;

If(Matcher.find()) then

{$C_1$, $C_2$…….. $C_n$ } add (Matcher.group()) _ R

{R _ VBR}

Else

{R _ NR}

NR _ {Case Marker Rules};

**Output:**

{W$_i$} = {R}+ {VB/N Suffixes}

**Conditional Random Field**

The CRF techniques as modeling in the learning phase and inference in the classification. CRF uses the conditional probability *P (label sequence* **y** *| observation sequence* **x***)* rather than the joint probability *P*(**y, x**) as in case of HMM. It specifies the probability of possible label sequences **y** for a given observation sequence **x**. CRF allows arbitrary, non-independent features on **x** while HMM does not. Probability of transitions between labels may depend on past and future observations. This technique has two phases for clause boundary identification:

*1. Learning:* Given a sample set X containing features {X$_1$, ... ,X$_n$ } along with the set of values for hidden labels Y i.e. clause boundaries{Y$_1$ , ... ,Y$_n$ }, learn the best possible potential functions.

*2. Inference*: For a given word there is some new observable x, find the most likely clause boundary y* for x, i.e. compute (exactly or approximately).

In this CRF technique linguistic rules are used as features for which different length of windows, comprises of words, are formed that depend on these linguistic rules. For example, in case of relative clause identification in Tamil language, clause beginning and ending are identified via rule1 and rule2 respectively.

**RULE_1:**

If the current word is any relative clause marker and next word is any of the POS tags verb, pronoun, adjective, noun then the next word is marked as beginning of clause boundary as shown below

*Position 0: Relative clause marker*
*Position 1: Verb or Adjective or noun or pronoun*

Then 0 should be marked as beginning of subordinate clause of type relative. Where position 0 indicates the current word and position 1 is the next word.

**RULE_2:**

If the current word is any verb auxiliary and next word is any symbol then current word is end of corresponding subordinate clause boundary as shown below

*Position 0: Verb phrase or Verb auxiliary*
*Position 1: any symbol or phrase*

Then 0 should be marked as end of above subordinate clause.

## V. RESULTS AND DISCUSSION

The Proposed frame work and algorithm is experimented with 150 sentences of text from the news papers and articles. All the sentences are used for training. In order to evaluate the system, we applied 150 test sentences, in that 120 sentences are correctly parsed. Moreover, Precision and Recall of words are widely used metrics to evaluate the efficiency of dependency relation between the words in the sentences. Precision is the percentage of generated words that are actually correct. The recall stands for the percentage of words that are generated and that are actually found in the reference translation. F-Measure is the harmonic mean of recall and precision.

**Table 1. Relationship Analysis**

| Input | Without dependency relation(MT) | With dependency relation(MT) |
|---|---|---|
| ☐☐☐☐ ☐☐☐☐☐☐☐☐☐ ☐☐☐☐☐☐☐☐. | He temple to went | He went to temple. |
| ☐☐☐☐ ☐☐☐☐☐☐☐☐ ☐☐☐☐☐ ☐☐☐☐☐☐☐☐☐☐ ☐☐☐☐☐☐☐☐. | He English for ten years studied | He studied English for ten years |
| பசு பால் கொடுக்கும் | The Cow milk gives | The cow gives milk. |
| நான் ஒரு பையனைக் கூடத்தில் பார்த்தேன் | I a boy in hall saw. | I saw a boy in the hall. |
| பம்பாய் இந்தியாவின் நுழைவாயில் | Bombay india gateway. | Bombay is the gateway of india. |
| அது ஓர் அழகான குழந்தை | It is a beautiful baby. | It is a beautiful baby. |

**Table 2. Experimental analysis of various metrics**

| Relation between the words | *P(%) | *R(%) | *F(%) |
|---|---|---|---|
| Without dependency | 63.78 | 56.32 | 59.82 |
| With dependency | 82.78 | 93.67 | 87.89 |

**\*P-Precision, \*R-Recall, \*F-F-Measure**

The comparative study is made with Dependency relation. The same set of data are used with dependency relation, out of which 120 sentences are correct, the main reason is semantic analyzed of prepositions and reordering error. We obtained the precision, recall and F-measure as 82.78 %, 93.67 %, 87.89 % respectively is as shown in table 2. The Fig. 8 emphasizes very clearly that the proposed system performance is better with respect to all the metric.

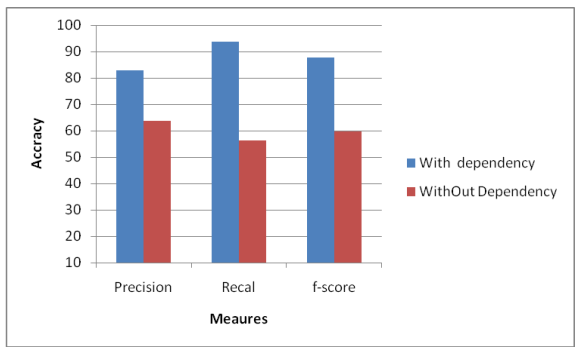

**Fig 8. Comparative study of dependency relation**

## VI. CONCLUSION

In this paper Conditional Random Fields are used for classification of clause boundary beginning and ending and also identify the type of subordinate clause. Limitation with CRFs is that it is highly dependent on linguistic rules. Missing of these rules may lead to wrongly classified data. An improvement can be achieved in this proposed clause markers for different subordinate clauses and also for those clauses which are embedded in the main clause. The proposed model will give good accuracy for all the sentences if the training data is increased. More complex sentence structures can be added to the training data. Also, the sentence in passive voice can be added to get good results.